\documentclass[letterpaper, 10 pt, conference]{ieeeconf}  

\IEEEoverridecommandlockouts                              

\usepackage{hyperref} 
\usepackage{mathrsfs}
\usepackage{amsfonts}
\usepackage{amsmath}
\usepackage[flushleft]{threeparttable}
\usepackage{tablefootnote}
\usepackage{multirow}
\usepackage{graphicx}
\usepackage{subfigure}
\let\labelindent\relax
\usepackage{enumitem}
\usepackage{hanging}
\usepackage{color}
\usepackage{tikz}
\usetikzlibrary{calc,arrows,decorations.markings}
\usepackage{balance}
\usepackage{cite} 

\graphicspath{{../},{../image}}

\usepackage{booktabs}  
\usepackage{colortbl}  

\usepackage{cuted}
\usepackage{capt-of}

\title{\LARGE \bf OmniPose6D: Towards Short-Term Object Pose
Tracking in Dynamic Scenes from Monocular RGB}

\author{Yunzhi Lin$^{1,2}$,  Yipu Zhao$^{1}$, Fu-Jen Chu$^{1}$, Xingyu Chen$^{1}$, Weiyao Wang$^{1}$, \\
Hao Tang$^{1}$, Patricio A.  Vela$^{2}$, 
Matt Feiszli$^{1}$, Kevin Liang$^{1}$ 
\\ $^{1}$Meta: {\tt\small \{yipuzhao, xingyuchen, fujenchu\}@fb.com};
\\ {\tt\small \{weiyaowang, haotang, mdf, kevinjliang\}@meta.com}
\\ $^{2}$Georgia Institute of Technology: {\tt\small \{yunzhi.lin, pvela\}@gatech.edu}
\thanks{Work was completed while the first author was an intern at Meta.}%
\thanks{This work was supported in part by NSF Award \#2026611.}%
}

\usepackage{xspace}
\makeatletter
\DeclareRobustCommand\onedot{\futurelet\@let@token\@onedot}
\def\@onedot{\ifx\@let@token.\else.\null\fi\xspace}
\def\eg{\emph{e.g}\onedot}

\usepackage{titlesec}

\begin{document}

\maketitle
\thispagestyle{empty}
\pagestyle{empty}


\begin{abstract}
To address the challenge of short-term object pose tracking in dynamic environments with monocular RGB input, we introduce a large-scale synthetic dataset OmniPose6D, crafted to mirror the diversity of real-world conditions. We additionally present a benchmarking framework for a comprehensive comparison of pose tracking algorithms. We propose a pipeline featuring an uncertainty-aware keypoint refinement network, employing probabilistic modeling to refine pose estimation. Comparative evaluations demonstrate that our approach achieves performance superior to existing baselines on real datasets, underscoring the effectiveness of our synthetic dataset and refinement technique in enhancing tracking precision in dynamic contexts. Our contributions set a new precedent for the development and assessment of object pose tracking methodologies in complex scenes.

\end{abstract}

\section{Introduction}
\label{sec:intro}

Predicting an object's 3D position and orientation in relation to a camera over time, known as 6-DoF object pose tracking, is crucial for applications such as robotic manipulation and augmented reality. This technology facilitates seamless interaction between robots and their targets, or allows digital content to be accurately overlaid onto the real world, enhancing a user's immersive experience.

Prior efforts in this domain have predominantly been model-based, relying on CAD models~\cite{pauwels2015simtrack, tjaden2017real, deng2021poserbpf}, pre-determined categories~\cite{wang20206, weng2021captra, lin2022keypoint}, or registration at inference time~\cite{yen2021inerf, labbe2023megapose, liu2022gen6d, he2022onepose++}. These assumptions inherently limit the ability of these methods to scale or adapt to the range of objects, trajectories, and environments observed in more diverse settings (\eg Ego4D~\cite{grauman2022ego4d}). More recently, model-free approaches have shown promise, delivering noteworthy accuracy and speed. Nonetheless, these methods~\cite{wen2021bundletrack, wen2023bundlesdf} are dependent on depth information, which requires additional sensors and restricts their utility for objects with challenging material properties. 
An RGB-only monocular method for object pose tracking would offer broad applicability, yet such methods~\cite{du2022pizza} are notably scarce and face great limitations, especially in dynamic environments and non-static scenes.

\begin{figure}[t]
  \vspace{-4mm}
    \centering
    \includegraphics[width=0.9\linewidth]{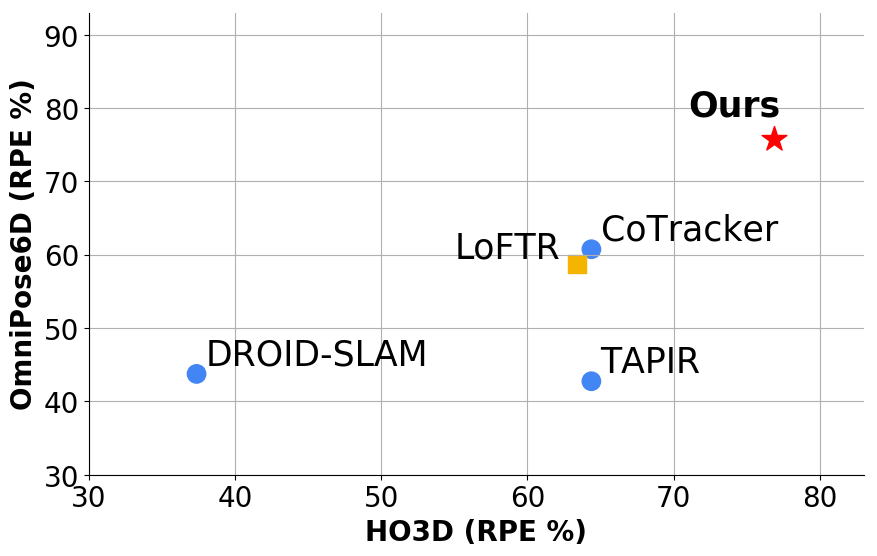}
    \vspace{-3mm}
  \caption{{\bf Object Pose Tracking Performance.} Our uncertainty-aware keypoint refinement approach (red star) outperforms baselines on both OmniPose6D (synthetic) and HO3D~\cite{hampali2020honnotate} (real) in Relative Pose Error (RPE) under 3 degrees. Blue dots and yellow square denote tracking and local feature matching methods, respectively.}
  \label{fig:teaser}
  \vspace{-6mm}
\end{figure}

Addressing these challenges, we present a monocular RGB-based object pose tracking method designed for dynamic environments. We focus on a short-term window of a few keyframes, as opposed to a long-term one spanning minutes, often operating with a sliding window. While long-term tracking may prioritize smoothness across windows for continuity over time, our short-term setting emphasizes immediate precision within each window---important for applications requiring accurate tracking, while also establishing a foundation that future long-term methods can build upon. Our method assumes the object's rigidity and the availability of a 2D mask in the video's initial frame, and without any other prior knowledge of the object. We track object pose change via keypoint representation. This approach is independent of model-specific information, facilitating seamless incorporation into existing pose computation workflows.

We build upon prior track-any-point works~\cite{harley2022particle, doersch2022tapvid, doersch2023tapir, karaev2023cotracker} to bridge keypoint tracking and object pose estimation. While keypoint tracking focuses on translational or unidirectional motion, pose tracking accounts for broader motion patterns. Track-any-point methods emphasize long-term single-point tracking, whereas pose tracking relies on short-term point pairs for pose estimation. Our pipeline integrates advances in video segmentation~\cite{cheng2022xmem, kirillov2023segany, yang2023track}, using masked object images as input. While segmentation quality may vary, our approach remains robust, as the pose tracking pipeline is driven by keypoint consistency rather than strict segmentation accuracy. We initialize grid sample tracks via an off-the-shelf keypoint tracker~\cite{karaev2023cotracker} and refine them with an uncertainty-aware network, selecting reliable tracks for accurate pose estimation via structure-from-motion.

A major challenge in 6-DoF pose estimation is the difficulty of large-scale real-world annotations. Synthetic datasets~\cite{harley2022particle, greff2021kubric} ease label generation but often lack realistic out-of-plane rotation and occlusion. To address this, we introduce \textbf{OmniPose6D}, a large-scale synthetic dataset emphasizing object-centric diversity and realistic motion trajectories. Our photo-realistic renderings and varied motion patterns enhance sim-to-real transfer, improving generalization to real-world scenarios.

\begin{figure}[t!]
  \centering
    \includegraphics[width=\columnwidth]{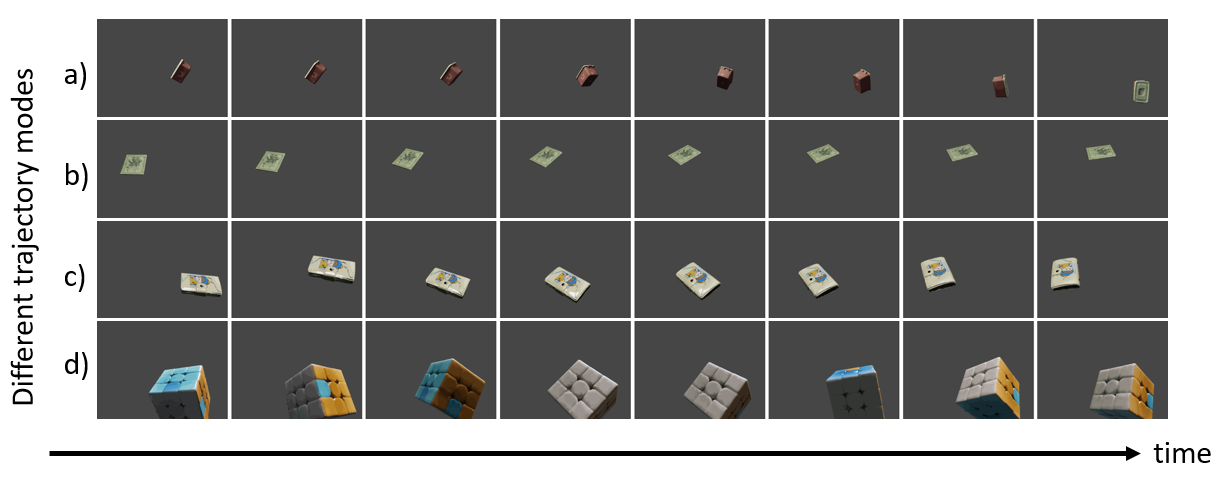}
    \vspace{-7mm}
  \caption{Sample sequences of trajectory modes in our dataset: a) random walk; b) circling camera; c) noisy trajectory; d) real dataset trajectory.}
  \label{fig:dataset_gen_traj}
  \vspace*{-6mm}
\end{figure}

We perform experiments on both a held-out portion of our OmniPose6D dataset and HO3D~\cite{hampali2020honnotate} real-world dataset. We compare against a diverse set of baselines, representing recent state-of-the-art methods from related problem settings, including local feature matching, simultaneous localization and mapping (SLAM), and keypoint tracking (Figure~\ref{fig:teaser}), combined with a Structure-from-Motion (SfM) framework for computing pose.
We observe that short-term RGB monocular pose tracking challenges baseline methods, while our synthetic training and uncertainty-aware pipeline achieve meaningful improvements.


Our contributions can be summarized as follows:

\begin{itemize}
\item We introduce RGB-only model-free object pose tracking in dynamic environments as a task and devise a simple yet effective pipeline with uncertainty-aware keypoint refinement, ensuring robust pose estimation.
\item We generate a large-scale synthetic dataset with 40K sequences of 100 frames each called {\bf OmniPose6D}. It contains a wide array of object meshes and motion trajectories, specifically curated for object pose tracking training and evaluation.
\item We demonstrate that the proposed method enhances generalization and robustness through extensive benchmarks in both synthetic and real-world scenarios.
\end{itemize}

\section{Related Work}
\label{sec:related}
\subsection{Object Pose Tracking}

Continuously estimating an object’s 3D position and orientation remains a core challenge in computer vision and robotics. Early instance-level pose tracking relied on filtering or least-squares optimization~\cite{choi2013rgb, wuthrich2013probabilistic, issac2016depth, deng2021poserbpf, pauwels2015simtrack, joseph2015versatile, tjaden2017real, li2018deepim, wen2020se}, while category-level methods used CAD models for training and template matching~\cite{lin2022keypoint, zhang2023genpose}, later evolving to feature-based matching for unseen objects~\cite{sun2022onepose, he2022onepose++, liu2022gen6d, labbe2023megapose, lin2023parallel}. Recent SLAM-inspired approaches~\cite{wen2021bundletrack, wen2023bundlesdf} enable online adaptation but still rely on depth.
While RGB-D methods show promise, monocular RGB-only, model-free techniques remain underexplored. 
While RGB-based method PIZZA~\cite{du2022pizza} employs a transformer-based framework that processes three consecutive frames for 6D pose estimation, this fixed temporal window can render its predictions sensitive to large inter-frame motions, thereby limiting robustness in highly dynamic environments. In contrast, our approach extends RGB-only methods by concentrating on short-term object pose tracking, which better accommodates abrupt motion changes and complex scene dynamics. Inspired by recent advances in 2D tracking~\cite{valmadre2018long}, our method integrates adaptive mechanisms designed to enhance reliability and versatility.

\vspace{-1mm}
\subsection{Keypoint Tracking}


Recent advances in keypoint tracking have improved robustness. Particle Video Revisited enhances tracking through occlusions via a sliding window~\cite{harley2022particle}, while TAP-Vid benchmarks physical point tracking~\cite{doersch2022tapvid}. TAPIR refines this by integrating TAP-Vid’s matching with PIPs’ occlusion handling~\cite{doersch2023tapir}.
CoTracker takes a different approach, using dense optical flow and transformers to track point groups~\cite{karaev2023cotracker}. We adapt its framework for keypoint refinement, focusing on precise tracking rather than full object pose estimation. With video segmentation reducing the need for extensive point visibility analysis~\cite{yang2023track}, our method prioritizes selecting well-tracked points through uncertainty estimation, improving pose tracking accuracy.

\subsection{Datasets}


Synthetic keypoint tracking datasets are favored for their precise ground truth labels. FlyingThings simulates object motion with physics engines but lacks photorealism, causing a sim-to-real gap~\cite{mayer2016large, harley2022particle}. TAP-Vid-Kubric improves realism using Blender rendering~\cite{greff2021kubric}, while MegaDepth provides real-world data, deriving keypoints from depth and camera poses~\cite{li2018megadepth}.
However, existing datasets lack dynamic object pose tracking, especially for interactions and diverse rotations. To address this, we introduce OmniPose6D, a synthetic dataset tailored for training and evaluating pose tracking in interactive settings, fostering future research in dynamic object pose tracking.

\begin{figure}[t!]
  \centering
    \includegraphics[width=\columnwidth]{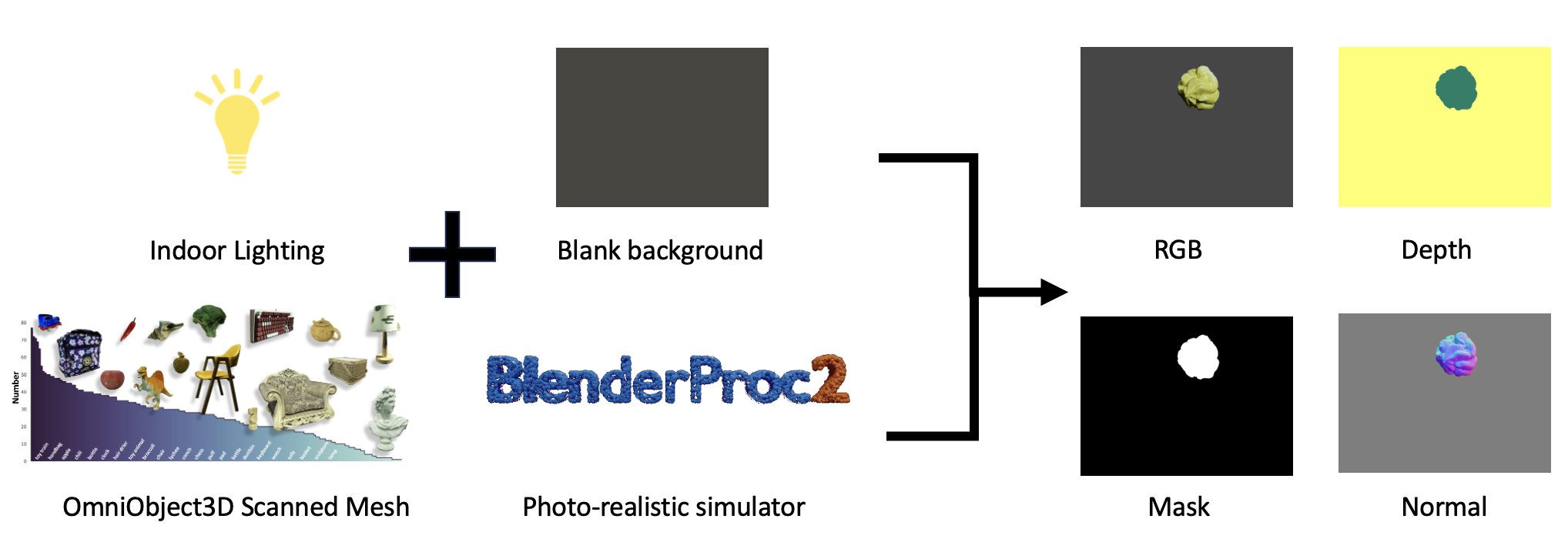}
  \caption{We utilize a large-scale scanned mesh dataset, OmniObject3D~\cite{wu2023omniobject3d}, combined with indoor lighting and a blank background. We efficiently simulate complex object movement trajectories with BlenderProc~\cite{Denninger2023}, generating comprehensive RGB, depth, mask, and normal annotations.} 
  \label{fig:dataset_gen_process}
  \vspace*{-5mm}
\end{figure}

\begin{figure*}[t]
  \vspace{1 mm}
  \centering
    \includegraphics[width=0.9\textwidth]{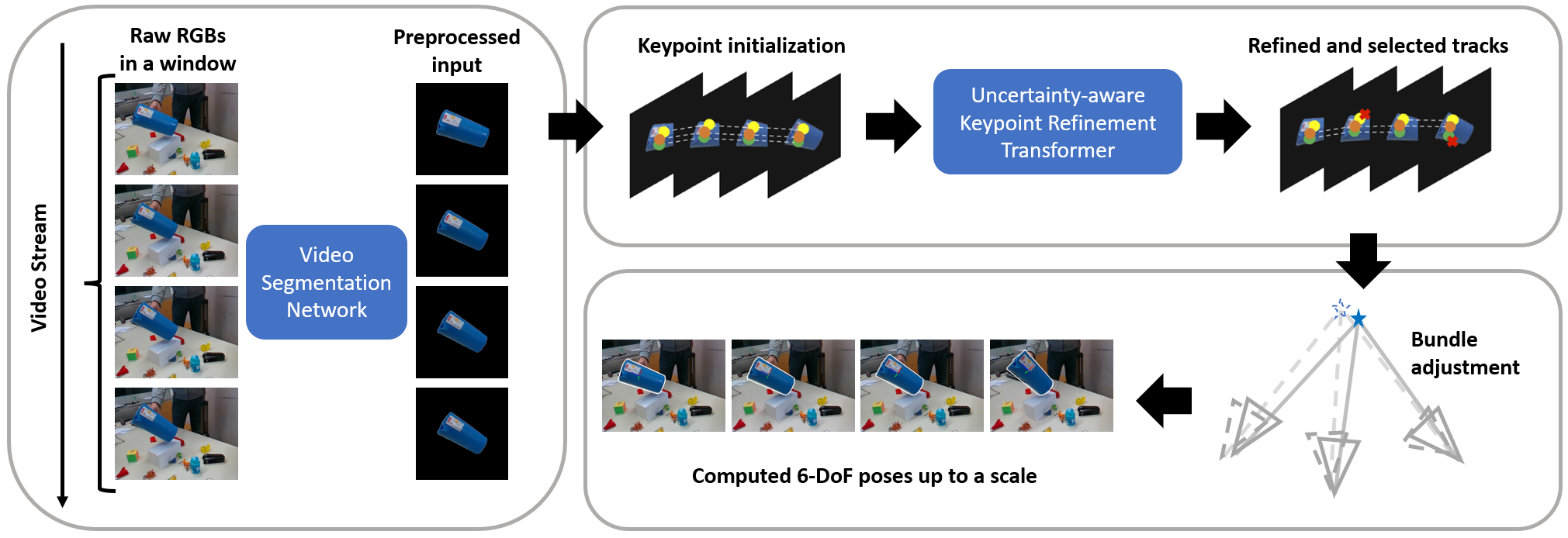}
  \caption{\textbf{Proposed pipeline.} Given a segmentation mask in the first frame, the object is tracked and cropped from a sequence of monocular RGB images with video segmentation~\cite{yang2023track} (Example images are from HO3D~\cite{hampali2020honnotate}).
  Rather than relying on precise segmentation, we perform dense grid sampling on the first mask to initialize keypoint queries for CoTracker~\cite{karaev2023cotracker}, enabling robust tracking of consistent keypoints. An uncertainty-aware transformer further refines and selects reliable tracks, ensuring stable pose estimation even with imperfect segmentation. The selected keypoints are then used to compute 6-DoF poses within a Structure-from-Motion framework. 
    \label{fig:pipeline}}
  \vspace*{-5mm}
  \end{figure*}
  
\section{OmniPose6D}
\label{sec:dataset}

\noindent {\bf Motivation:} The lack of large-scale, precise ground truth data hampers the development of dynamic object pose tracking research. Existing real datasets~\cite{hampali2020honnotate, chao2021dexycb, fan2023arctic} focus exclusively on hands and lack the variety of objects and accuracy of keypoint annotations. Meanwhile, synthetic datasets~\cite{mayer2016large, harley2022particle, greff2021kubric} provide rich keypoint labels but overlook object-specific dynamics and rotational movements. To overcome those limitations, we propose a new, expansive photo-realistic synthetic dataset, OmniPose6D, that unites detailed keypoints with an object-centric perspective, crafted for high-precision, dynamic tracking.

\noindent {\bf Simulator Choice:} We use BlenderProc~\cite{Denninger2023} for its photorealism, user-friendly Python scripting, and support for BOP-format output, ensuring our generated annotations (RGB, depth, pose, normals, segmentation) are readily usable.

\noindent {\bf Mesh Choice:} Instead of using ShapeNet~\cite{chang2015shapenet} or YCB~\cite{singh2014bigbird} alone, we leverage well-scanned, diverse meshes from OmniObject3D~\cite{wu2023omniobject3d}. We optimize these meshes with Blender~\cite{blender} to reduce size and remove invalid ones, resulting in over 4,000 meshes across more than 200 categories.



\noindent {\bf Motion Generation:}
Capturing diverse object trajectories is crucial to pose tracking generalization. Unlike other keypoint tracking datasets~\cite{harley2022particle, greff2021kubric}, which initialize objects once and rely solely on physics engines for motion simulation, we simulate complete trajectories with four distinct modes:
\begin{itemize}
\item \textit{Random Walk}: Manipulated objects can have significantly variable trajectories, which we model as a random walk in 3D space with a fixed camera.
\item \textit{Circling Camera}: Objects are often observed from multiple angles at ground level, so we simulate a camera circling at varying heights and viewpoints.
\item \textit{Noisy Trajectory}: To mimic real-world unpredictability, we introduce noise into the circling camera trajectory, creating a more erratic motion path.
\item \textit{Real Dataset Trajectories}: To reduce the sim2real gap, we incorporate trajectories from the training split of real hand-object interactions~\cite{hampali2020honnotate}. Longer sequences can be chopped into multiple shorter ones and applied to different objects, significantly increasing variability.
\end{itemize}
Through these modes, we aim to more accurately reflect the complexities of real-world object motion, enhancing the efficacy of object pose tracking systems.

\noindent {\bf Background Choice:}
While many datasets incorporate indoor and outdoor scenes as backgrounds, we choose to use a blank background to focus on the foreground objects. Our pipeline works with segmented regions, reducing the need for predefined backgrounds. In the meantime, we provide depth and instance segmentation labels to offer flexibility, enabling users to easily generate and integrate their own backgrounds during post-processing, if needed.

\noindent {\bf Summary:} We have generated a large-scale photo-realistic dataset (OmniPose6D) of 40,000 sequences, each comprising 100 frames, distributed across 203 distinct categories, of which 21 are reserved for validation. OmniPose6D incorporates four trajectory modes, each encompassing 10,000 sequences and featuring a single object, exhibiting varied movements.
To the best of our knowledge, this dataset represents the first and most comprehensive resource for object pose tracking within dynamic settings.
The sequences, produced through our pipeline, boast photo-realistic quality and a wide range of variation, thereby facilitating enhanced generalization capabilities compared to existing datasets.

\begin{figure*}[t!]
  \centering
  \vspace{1mm}

  \begin{minipage}{0.6\textwidth}
    \centering
    \includegraphics[width=\textwidth]{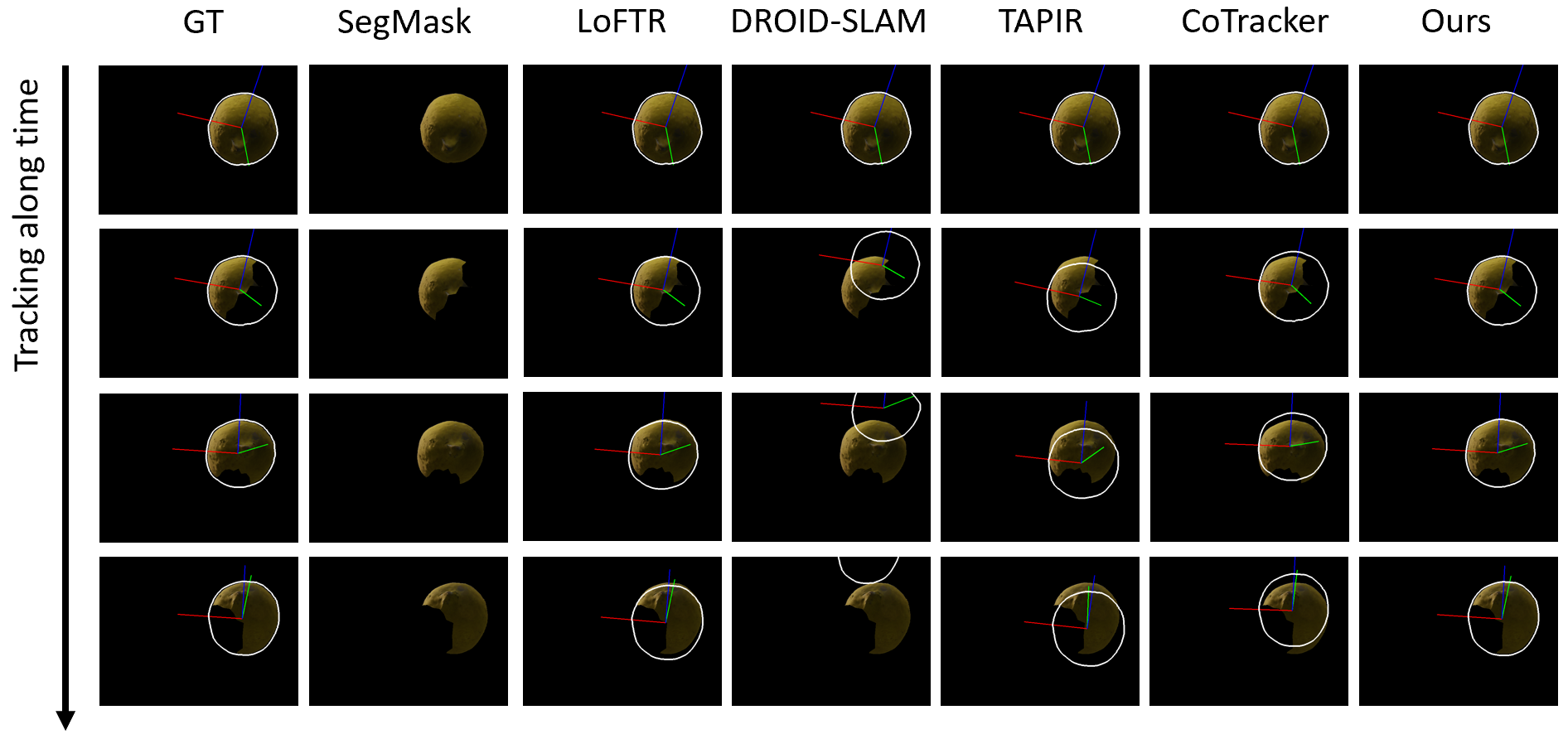}
    \vspace{-6mm} \\ 
    {\small (a) OmniPose6D dataset}
  \end{minipage}

  \vspace{1mm}

  \begin{minipage}{0.6\textwidth}
    \centering
    \includegraphics[width=\textwidth]{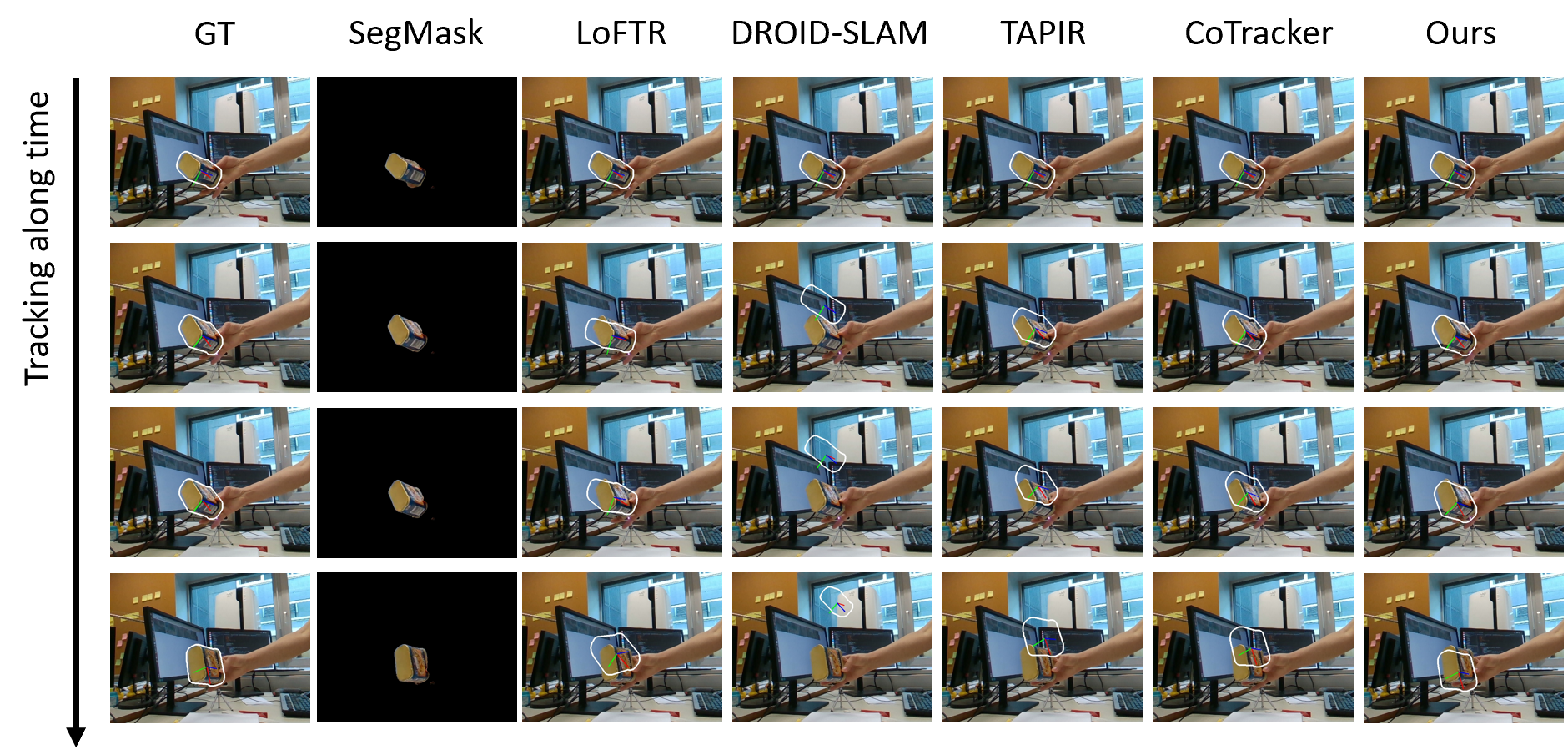}
    \vspace{-6mm} \\
    {\small (b) HO3D dataset}
  \end{minipage}


  \caption{Qualitative comparison of the methods on two datasets: (a) OmniPose6D dataset and (b) HO3D~\cite{hampali2020honnotate} dataset. In both images, we visualize 6-DoF pose tracking results, where the contour (white) is rendered with the estimated pose. Our method tracks the object's pose successfully in both cases, while other methods perform differently. We also demonstrate that our proposed pipeline remains effective even with imprecise segmentation (See the details of SegMask prediction of HO3D).}
  \label{fig:qualitative_comparison_combined}
  \vspace*{-4mm}
\end{figure*}

\section{Uncertainty-aware Keypoint Refinement}
\label{sec:approach}

Given a monocular RGB input video and a segmentation mask that highlights the object of interest in the first frame, our method tracks the 6-DoF pose of the object relative to the first frame through subsequent frames, up to a scale factor (overview in Figure~\ref{fig:pipeline}). 
Our method operates under the sole assumption of the object's rigidity. It does not depend on any further assumptions, such as the availability of an instance-level CAD model during training, category-level priors, or a model-scan phase during the testing period.
Note that the segmentation operation identifies a rough area of the target but inevitably introduces some flaws. We do not assume perfect segmentation is available.

\subsection{Preliminaries}

Our approach is to refine the keypoint tracks and estimate the uncertainty associated with each keypoint. To achieve this, we utilize the foundational design of CoTracker~\cite{karaev2023cotracker}, which is adept at tracking groups of points and discerning their correlations. This approach is especially relevant to our work, as we focus on keypoints derived from a single object, exhibiting intrinsic correlations.
The network processes a sequence of \(T\) RGB frames \(I_t \in \mathbb{R}^{3 \times H \times W}\), creating a track \(P^i = (x^i_t, y^i_t) \in \mathbb{R}^2\) for each of the \(N\) points across a timeframe \(t = t^i, \ldots, T\), with \(t^i\) marking the track's start. 
It also predicts \textit{visibility flag} \(v^i_t \in \{0, 1\}\), indicating the visibility or occlusion of points in each frame.
We empirically set $T=8$ for our short-term window, balancing computational efficiency and effective downstream SfM performance.
We harness CoTracker's robust methods of feature extraction, including image, track, and correlation features, along with the application of positional encoding for both keypoints and time frames. Utilizing a transformer architecture with interleaving time and group attention blocks, the network updates the tracks iteratively. 
Our implementation capitalizes on the established techniques of the CoTracker network, adapting them to align with requirements inherent in our pursuit of keypoint tracking and uncertainty estimation.

\subsection{Training Data Generation}
\label{sec:training}

To enhance frame rate diversity, we initially sample $N$ frames sequentially from the original sequence, ensuring that each interval does not exceed a maximum distance of $K$ frames; we empirically set $N=8$ and $K=4$ in our experiments. This ensures variability in the training data.
Object poses are intriguing during hand interaction, often resulting in substantial occlusions. Therefore, we simulate occlusions along the mask boundaries in selected images to mimic these conditions. Specifically,  We draw perturbed black circles centered on random points along the mask boundary; their radiuses are proportional ([0.1, 0.5]) to the size of the mask.
To ensure adequate useful keypoints in the processed inputs, we employ a co-visibility approach inspired by~\cite{he2023dfsfm}.
For establishing keypoint correspondences, the first view is designated as the reference. We project grid-level points sampled on the reference view's mask to other views. The projection is deduced from the provided depth maps and pose labels.
We also compute the visibility of each keypoint by using the projected depth error and cycle projection error mentioned in~\cite{he2023dfsfm}.
Finally, the input is enhanced with several image augmentation operations, including color jitter, Gaussian blur, motion blur, additive shade, and random flipping.
We apply rough segmentation to identify the small region of interest within the image and crop it accordingly. The cropping parameters, centered on the segmented area, are adjusted with random shifts and scaling to ensure robustness. This process standardizes the input and the affine transformation matrix is recorded to map the crop back to the original image context.

\subsection{Uncertainty Estimation}

To estimate keypoint uncertainty in a probabilistic framework, we reformulate the prediction of the keypoint error as a classification problem with $n_L$ predefined levels. For each keypoint $i$, let the true error $e_i$ be the Euclidean (L2) distance between its predicted location and the ground truth. We discretize the range of possible errors into $n_L$ intervals defined by thresholds 
\[
L = [l_1, l_2, \dots, l_{n_L}],
\]
where each $l_k$ represents the $k$-th error boundary. We empirically set $n_L = 5$ with the thresholds 
\[
L = [1, 3, 5, 10, \infty],
\]
which are adjusted based on the ratio between the input image size at inference and the training image size to ensure consistent performance across resolutions. The ground truth label $y_i$ for keypoint $i$ is determined as follows ($l_0=0$):
\[
y_i =
\begin{cases}
k, & \text{if } l_{k-1} \le e_i < l_k \quad \text{for } k = 1,\dots,n_L-1, \\
n_L, & \text{if } e_i \ge l_{n_L},
\end{cases}
\]

Our network outputs a set of logits $\{z_{i,k}\}_{k=1}^{n_L}$ for each keypoint $i$, which are converted into probabilities via the softmax function:
\[
p_{i,k} = \frac{\exp(z_{i,k})}{\sum_{j=1}^{n_L} \exp(z_{i,j})}, \quad k = 1,\dots,n_L.
\]
Here, $p_{i,k}$ represents the probability that keypoint $i$’s error falls into the $k$-th error interval.

To train the network, we use a balanced cross-entropy loss between the predicted distribution and the ground truth labels:
\[
\mathcal{L}_{\text{uncertainty}} = - \sum_{i} \sum_{k=1}^{n_L} \mathbb{I}(y_i = k) \log p_{i,k},
\]
where $\mathbb{I}(\cdot)$ is the indicator function.

This formulation allows the network to learn a discrete probability distribution over the error levels, thus providing a probabilistic interpretation of the keypoint uncertainty. The resulting uncertainty estimates are then used to rank and select keypoints, ensuring that only the most reliable ones contribute to the pose estimation process.

\subsection{Object Pose Estimation}

Our objective is to determine the 6-DoF pose relative to a scale through tracked keypoints. The process begins by applying the inverse of the stored affine transformation to the predicted keypoints, converting them back to their initial image input coordinates. Subsequently, we apply the visibility prediction to exclude occluded keypoints.
We next apply a softmax function to the uncertainty estimates, identifying the most probable uncertainty class associated with each keypoint. We proceed by sorting the keypoints within each view according to their uncertainty class label, retaining the top $R$ ratio ($R=0.95$ in our paper) with the lowest associated uncertainty.
Keypoints are then organized into exhaustive view pairs, maintaining only the pairs where the corresponding keypoint is valid in both views. Finally, we employ COLMAP~\cite{schonberger2016structure}, a robust Structure-from-Motion framework, to perform bundle adjustment and calculate 6-DoF pose for each view within the window.

\begin{table*}[t]
\vspace{2mm}
\caption{Quantitative evaluation of OmniPose6D and HO3D datasets with KITTI Odometry metrics. We report average precision ($\uparrow$), conditional on error below several thresholds.}
  \centering
  \begin{threeparttable}
  \resizebox{0.8\linewidth}{!}{
    \begin{tabular}{cccccccccccccccc}
      \toprule
      \multirow{2}{*}{Method} & \multicolumn{3}{c}{$t_{err}$ (\%)} & \multicolumn{3}{c}{$r_{err}$ (deg/m)} & \multicolumn{3}{c}{ATE (m)} & \multicolumn{3}{c}{RPE (m)} & \multicolumn{3}{c}{RPE (deg)} \\
      \cmidrule(r){2-4} \cmidrule(r){5-7} \cmidrule(r){8-10} \cmidrule(r){11-13} \cmidrule(r){14-16}
                            & 10 & 20 & 30 & 30 & 40 & 50 & 0.01 & 0.03 & 0.05 & 0.01 & 0.015 & 0.025 & 1 & 2 & 3 \\
      \midrule
      \multicolumn{16}{c}{\textbf{OmniPose6D Dataset}} \\
      \midrule
      GT                & 0.62 & 0.71 & 0.76 & 0.69 & 0.73 & 0.76 & 0.78 & 0.84 & 0.86 & 0.82 & 0.85 & 0.86 & 0.84 & 0.89 & 0.90 \\
      LoFTR~\cite{sun2021loftr} & \underline{0.20} & \underline{0.30} & \underline{0.36} & \underline{0.14} & \underline{0.19} & \underline{0.26} & \underline{0.44} & 0.63 & 0.68 & 0.57 & 0.63 & 0.70 & \underline{0.34}& 0.52  & 0.60 \\
      DROID-SLAM~\cite{teed2021droid} & 0.06 & 0.11 & 0.14 & 0.10 & 0.13 & 0.15 & 0.21 & 0.50 & 0.62 & 0.44 & 0.57 & 0.68 & 0.17 & 0.36 & 0.45 \\
      TAPIR~\cite{doersch2023tapir} & 0.03 & 0.13 & 0.21 & 0.07 & 0.09 & 0.12 & 0.31 & 0.62 & 0.71 & 0.53 & 0.64 & 0.73 & 0.10 & 0.32 & 0.44 \\
      CoTracker~\cite{karaev2023cotracker} & 0.12 & 0.25 & 0.34 & 0.11 & 0.18 & 0.22 & 0.43 & \underline{0.64} & \underline{0.73} & \underline{0.61} & \underline{0.69} & \underline{0.76} & 0.33 & \underline{0.52} & \underline{0.62} \\
      Ours                & \textbf{0.34} & \textbf{0.51} & \textbf{0.59} & \textbf{0.35} & \textbf{0.42} & \textbf{0.48} & \textbf{0.62} & \textbf{0.74} & \textbf{0.80} & \textbf{0.71} & \textbf{0.76} & \textbf{0.80} & \textbf{0.62} & \textbf{0.71} & \textbf{0.74} \\
      \midrule
      \multicolumn{16}{c}{\textbf{HO3D Dataset}} \\
      \midrule
      GT                & 0.96 & 0.97 & 0.97 & 0.96 & 0.97 & 0.97 & 0.96 & 0.97 & 0.98 & 0.97 & 0.97 & 0.98 & 0.97 & 0.97 & 0.98 \\
      LoFTR~\cite{sun2021loftr} & 0.01 & 0.13 & 0.31 & 0.10 & 0.19 & 0.27 & 0.10 & 0.46 & 0.64 & 0.27 & 0.47 & 0.65 & 0.07 & 0.43 & 0.64 \\
      DROID-SLAM~\cite{teed2021droid} & 0.00 & 0.06 & 0.14 & 0.05 & 0.08 & 0.11 & 0.06 & 0.26 & 0.33 & 0.17 & 0.28 & 0.37 & 0.08 & 0.26 & 0.38 \\
      TAPIR~\cite{doersch2023tapir} & 0.01 & 0.10 & 0.28 & 0.08 & 0.16 & 0.25 & 0.07 & 0.47 & 0.62 & 0.25 & 0.47 & 0.67 & 0.05 & 0.39 & 0.65 \\
      CoTracker~\cite{karaev2023cotracker} & \underline{0.01} & \underline{0.15} & \underline{0.35} & \underline{0.11} & \underline{0.19} & \underline{0.29} & \underline{0.15} & \underline{0.51} & \underline{0.65} & \underline{0.34} & \underline{0.52} & \underline{0.70} & \underline{0.15} & \underline{0.51} & \underline{0.65} \\
      Ours                & \textbf{0.02} & \textbf{0.22} & \textbf{0.46} & \textbf{0.19} & \textbf{0.32} & \textbf{0.44} & \textbf{0.17} & \textbf{0.64} & \textbf{0.78} & \textbf{0.39} & \textbf{0.62} & \textbf{0.79} & \textbf{0.18} & \textbf{0.61} & \textbf{0.76} \\
      \bottomrule
    \end{tabular}
  }
  \end{threeparttable}
\label{tab:combined_omnipose6d_ho3d}
 \vspace{-5mm}
\end{table*}

\section{Experiments}
\label{sec:exp}

In this section, we aim to demonstrate that our tracking algorithm achieves state-of-the-art performance on short-term object pose tracking in dynamic settings on both synthetic and real benchmarks. More details, including ablations and failure cases, are provided in the supplementary video.

\noindent {\bf Metrics:}
As we work on a novel challenge without a pre-existing, dedicated evaluation metric, we adopt universally recognized evaluation criteria to ensure a fair and accessible benchmark for future work, believing the KITTI Odometry metrics~\cite{zhan2019dfvo} fit our task most. 
While it was originally designed for camera pose evaluation, the metrics perfectly work for our case, as the pose of a single object is inherently defined by the inverse of its relative camera pose.
It encompasses several key metrics, including the average translational error ($t_{err}$ in \%) and rotational errors ($r_{err}$ in \%), which are evaluated over potential sub-sequences of varying proportions (0.1, 0.2, …, 1). It also includes Absolute Trajectory Error (ATE), which calculates the root-mean-square error between the predicted camera poses [x, y, z] and their ground truth counterparts, and the Relative Pose Error (RPE), which quantifies the frame-to-frame relative pose inaccuracies in terms of both translation and rotation.
Given our focus on benchmarking monocular RGB methods, which inherently lack a real-world scale reference, we adopt a 7-DoF (Degrees of Freedom) optimization technique~\cite{umeyama1991least} for scaling and alignment. This approach minimizes the ATE against the ground truth poses, addressing the absence of a scaling factor in our predictions.
Due to disproportionately large errors from the most difficult examples, we propose reporting the average precision for each metric conditioned on error below a specified threshold for a more balanced assessment, similar to~\cite{li2018deepim,Wang_2019_CVPR}. The specific threshold number is put under each metric in the table.

\noindent {\bf Baselines:}
We select representative works from related fields: LoFTR~\cite{sun2021loftr} (local feature matching), DROID-SLAM~\cite{teed2021droid} (simultaneous localization and mapping), and TAPIR~\cite{doersch2023tapir} and CoTracker~\cite{karaev2023cotracker} (keypoint tracking). 
We use the RGB-only version of DROID-SLAM. Except for DROID-SLAM, which can compute poses directly, other methods focus on either keypoint matching or tracking. For a fair comparison, we integrate these methods into a uniform Structure-from-Motion (SfM) framework~\cite{schonberger2016structure}, allowing consistent pose computation across all techniques.

\noindent {\bf Implementation details:}
We train our network with a batch size of 32 on 16 NVIDIA V100 GPUs for 100 epochs (5,000 samples per epoch), using a 5e-4 learning rate with a linear 1-cycle schedule and AdamW~\cite{loshchilov2017decoupled} optimizer. 
For inference, we densely sample keypoints on the initial mask with a 3-grid size and randomly select 1,500 tracks due to memory constraints. The input is cropped to 512×512 using segmentation masks. Processing an 8-frame input to estimate poses takes approximately 10 seconds in total on a single NVIDIA 3090 GPU. 
We adopt the mapper function in COLMAP with the default parameters to perform Structure-from-motion and compute the final poses.

\subsection{OmniPose6D Synthetic Dataset}
\label{exp:synthetic}

\noindent {\bf Setting.}
OmniPose6D has 21 distinct object categories for evaluation, each featuring a varying number of instances and sequences. To account for the long-tailed distribution, we randomly select 5 sequences from each category, with each sequence comprising 100 frames.
In these sequences, an object moves randomly in 3D space relative to the camera against a blank background. Since the original dataset lacks inter-object occlusion, we simulate occlusion randomly in each view (see Section~\ref{sec:training}). This process synthesizes potential occlusions caused by hands or other objects and includes imperfections in segmentation inputs. These images with corrupted masks are then used as inputs in our pipeline.
As our task concentrates on short-term window segments, we segment long sequences into shorter clips. To ensure each window has sufficient motion, we use a keypoint tracking method~\cite{karaev2023cotracker} to track the average displacement of randomly selected points, choosing keyframes where the motion surpasses a set threshold. The resulting set of keyframes are then evenly divided into multiple subsets. For our experiments, we set each window to encompass 8 frames. This choice is motivated by the need to balance accuracy and computational efficiency when processing keypoint tracking and bundle adjustment.

\noindent {\bf Results.}
Table~\ref{tab:combined_omnipose6d_ho3d} (top) shows quantitative results, and Figure~\ref{fig:qualitative_comparison_combined} (a) shows qualitative comparisons. Note we present the results using densely sampled {\it visible} Ground Truth (GT) keypoint correspondences, which is more accurate but still imperfect, due to object occlusions and errors introduced by the pose computation process (\eg coplanar points provide less geometric information).
Our method outperforms the baselines by a large margin on all the metrics.
LoFTR~\cite{sun2021loftr} excels in per-frame translation and rotation metrics due to its precise local feature matching, which is robust to random occlusions in individual frames. However, its lack of temporal consistency leads to accumulated errors over time, affecting ATE and RPE.
DROID-SLAM~\cite{teed2021droid} has more trouble with the test cases, as it focuses on efficiency more than accuracy, and the short-window setting does not favor its ability in large-scale scenes.
While TAPIR~\cite{doersch2023tapir} and CoTracker~\cite{karaev2023cotracker} are both keypoint tracking methods, CoTracker outperforms with its group point tracking, especially in tasks where points are correlated, such as object pose tracking.
Our proposed approach builds upon CoTracker for even better results. We attribute this to our comprehensive training dataset, which covers more complex scenes with more random occlusion for object pose tracking, and focuses on the most valuable keypoints to compute the poses, from uncertainty estimation.

\subsection{HO3D Real Dataset}
\label{exp:ho3d}

\noindent {\bf Setting.}
HO3D~\cite{hampali2020honnotate} features RGB-D videos of a human hand interacting with YCB objects, with ground truth derived from multi-view registration. In our experiments, we only utilize the RGB images, discarding depth. We employ the latest version, HO3D\_v3, and conduct tests on the official evaluation set, which comprises 4 distinct objects and 13 video sequences, each over 1000 images.
For efficiency, we first downsample the sequences by a factor of $M=3$. We then apply video segmentation~\cite{yang2023track} to obtain mask information throughout the sequence, using manual annotation of the first frame as a reference. While this segmentation method generally yields accurate results, the masks in later parts of the sequence may exhibit noise, particularly when the viewpoint significantly deviates from the initial frame. We still use these segmentations as our input to assess robustness to such variations. In alignment with the approach outlined in Section~\ref{exp:synthetic}, we employ the same strategy for dividing the long sequences into small subsets for analysis, setting each window to 8 frames.

\begin{table*}[t]
\vspace{1.5mm}
\caption{Impact of training dataset source on HO3D~\cite{hampali2020honnotate} object pose tracking.
}
  \centering
  \begin{threeparttable}
  \resizebox{0.75\linewidth}{!}{
\begin{tabular}{cccccccccccccccc}
\toprule
\multirow{2}{*}{Source} & \multicolumn{3}{c}{$t_{err}$ (\%)} & \multicolumn{3}{c}{$r_{err}$ (deg/m)} & \multicolumn{3}{c}{ATE (m)} & \multicolumn{3}{c}{RPE (m)} & \multicolumn{3}{c}{RPE (deg)} \\
\cmidrule(r){2-4} \cmidrule(r){5-7} \cmidrule(r){8-10} \cmidrule(r){11-13} \cmidrule(r){14-16}
                        & 10 & 20 & 30 & 30 & 40 & 50 & 0.01 & 0.03 & 0.05 & 0.01 & 0.015 & 0.025 & 1 & 2 & 3 \\
\midrule

Pre-trained & \underline{0.02} & \underline{0.18} & 0.36 & 0.12 & 0.24 & \underline{0.36} & 0.15 & 0.52 & 0.68 & \underline{0.34} & 0.54 & 0.72 & \underline{0.15} & 0.50 & 0.70 \\
Scratch     & 0.01 & 0.17 & \underline{0.38} & \underline{0.12} & \underline{0.25} & 0.35 & \underline{0.16} & \underline{0.60} & \underline{0.73} & 0.31 & \underline{0.54} & \underline{0.74} & 0.12 & \underline{0.53} & \underline{0.72} \\
Finetune    & \textbf{0.02} & \textbf{0.23} & \textbf{0.45} & \textbf{0.18} & \textbf{0.28} & \textbf{0.40} & \textbf{0.20} & \textbf{0.65} & \textbf{0.77} & \textbf{0.40} & \textbf{0.60} & \textbf{0.78} & \textbf{0.17} & \textbf{0.60} & \textbf{0.74} \\

\bottomrule
\end{tabular}
}
  \end{threeparttable}
\label{tab:data_source}
\end{table*}

\begin{table*}[t]
\caption{Input cropping is generally helpful for HO3D~\cite{hampali2020honnotate} object pose tracking.
}
  \centering
  \begin{threeparttable}
  \resizebox{0.75\linewidth}{!}{
\begin{tabular}{cccccccccccccccc}
\toprule
\multirow{2}{*}{Strategy} & \multicolumn{3}{c}{$t_{err}$ (\%)} & \multicolumn{3}{c}{$r_{err}$ (deg/m)} & \multicolumn{3}{c}{ATE (m)} & \multicolumn{3}{c}{RPE (m)} & \multicolumn{3}{c}{RPE (deg)} \\
\cmidrule(r){2-4} \cmidrule(r){5-7} \cmidrule(r){8-10} \cmidrule(r){11-13} \cmidrule(r){14-16}
                        & 10 & 20 & 30 & 30 & 40 & 50 & 0.01 & 0.03 & 0.05 & 0.01 & 0.015 & 0.025 & 1 & 2 & 3 \\
\midrule

w/o crop input                & 0.01 & 0.16 & 0.34 & 0.11 & 0.20 & 0.28 & 0.15 & 0.51 & 0.64 & 0.34 & 0.52 & 0.70 & 0.15 & \textbf{0.51} & 0.67 \\
w/ crop input                & \textbf{0.02} & \textbf{0.18} & \textbf{0.36} & \textbf{0.12} & \textbf{0.24} & \textbf{0.36} & \textbf{0.15} & \textbf{0.52} & \textbf{0.68} & \textbf{0.34} & \textbf{0.54} & \textbf{0.72} & \textbf{0.15} & 0.50 & \textbf{0.70}
 \\

\bottomrule
\end{tabular}
}
  \end{threeparttable}
\label{tab:crop}
\end{table*}

\begin{table*}[t]
\caption{Uncertainty estimation strategies: We test out several uncertainty estimation strategies for HO3D~\cite{hampali2020honnotate} object pose tracking.
}
  \centering
  \begin{threeparttable}
  \resizebox{0.75\linewidth}{!}{
\begin{tabular}{cccccccccccccccc}
\toprule
\multirow{2}{*}{Strategy} & \multicolumn{3}{c}{$t_{err}$ (\%)} & \multicolumn{3}{c}{$r_{err}$ (deg/m)} & \multicolumn{3}{c}{ATE (m)} & \multicolumn{3}{c}{RPE (m)} & \multicolumn{3}{c}{RPE (deg)} \\
\cmidrule(r){2-4} \cmidrule(r){5-7} \cmidrule(r){8-10} \cmidrule(r){11-13} \cmidrule(r){14-16}
                        & 10 & 20 & 30 & 30 & 40 & 50 & 0.01 & 0.03 & 0.05 & 0.01 & 0.015 & 0.025 & 1 & 2 & 3 \\
\midrule

w/o Uncertainty & 0.02 & \underline{0.23} & 0.45 & \underline{0.18} & 0.28 & 0.40 & \textbf{0.20} & \textbf{0.65} & \underline{0.77} & \textbf{0.40} & 0.60 & 0.78 & 0.17 & \underline{0.60} & 0.74 \\
by Level & \underline{0.02} & \textbf{0.23} & \underline{0.45} & 0.17 & \underline{0.31} & \underline{0.41} & \underline{0.18} & 0.62 & 0.76 & 0.38 & \underline{0.62} & \underline{0.78} & \underline{0.17} & 0.59 & \underline{0.75} \\
by Ranking & \textbf{0.02} & 0.22 & \textbf{0.46} & \textbf{0.19} & \textbf{0.32} & \textbf{0.44} & 0.17 & \underline{0.64} & \textbf{0.78} & \underline{0.39} & \textbf{0.62} & \textbf{0.79} & \textbf{0.18} & \textbf{0.61} & \textbf{0.76} \\
\bottomrule
\end{tabular}
}
  \end{threeparttable}
\label{tab:uncertainty}
\end{table*}

\noindent {\bf Results.}
Quantitative results are shown in Table~\ref{tab:combined_omnipose6d_ho3d} (bottom) and qualitative comparisons are illustrated in Figure~\ref{fig:qualitative_comparison_combined} (b). HO3D exhibits a higher degree of realism, due to the presence of image-level noise like variations in lighting and environmental factors, which may not be fully replicated in a synthetic dataset. The occlusion patterns from the hand are also different, which is more consistent compared to the higher occlusion pattern randomness of the synthetic dataset. 
Echoing Section~\ref{exp:synthetic}, our method again achieves the best performance. While the sim2real gap still exists, our large-scale synthetic dataset shows potential to help alleviate the problem.
We also observe that a consistent pattern emerges: methods that perform well on synthetic datasets tend to also excel on real-world datasets. This indicates that our custom-generated synthetic dataset serves as an effective proxy for evaluating different methodologies.

\subsection{Ablation Studies}
\label{sec:ablation}

We ablate on HO3D~\cite{hampali2020honnotate} for its more realistic settings.

\noindent {\bf Training Dataset Sources.}
%
We examine the impact of different training dataset sources on the performance of our system. 
We primarily focus on the comparison with CoTracker, which achieves the best overall performance across different test datasets.
To mitigate the influence of the uncertainty branch, 
we substitute the refinement network in Figure~\ref{fig:pipeline} with the original CoTracker’s~\cite{karaev2023cotracker} network but vary the training sources.
These include: 1) utilizing the original weights, which were trained on the TAP-Vid-Kubric dataset~\cite{doersch2022tapvid}; 2) training the network from scratch using our dataset; 3) using pre-trained weights and then fine-tuning them on OmniPose6D.
The results, as presented in Table~\ref{tab:data_source}, reveal that the combination of using pre-trained weights followed by fine-tuning on our dataset yields superior performance. Furthermore, training from scratch on our dataset also outperforms the approach of using the original weights. This indicates that OmniPose6D is both comprehensive and aptly suited for the task of object pose-tracking. Simultaneously, the TAP-Vid-Kubric~\cite{doersch2022tapvid} dataset effectively captures translational motion. Therefore, we recommend the combined use of both datasets to enhance coverage across wider scenarios.

\noindent {\bf Cropping Input Strategy.}
%
As outlined in Section~\ref{sec:training}, we preprocess the input with a crop on the mask. A comparison between the original CoTracker~\cite{karaev2023cotracker} with and without input cropping demonstrates that this operation, while straightforward, is notably effective (Table~\ref{tab:crop}).

\noindent {\bf Uncertainty Application Strategies.}
We aim to effectively leverage the estimated uncertainty of each keypoint. We consider two strategies: 1) Employing the predicted uncertainty to selectively discard keypoints (we remove points if their predicted levels lie in the most unreliable interval); 2) Organizing keypoints based on their uncertainty and preserving only those within the most reliable range. Our results, as detailed in the accompanying Table~\ref{tab:uncertainty}, reveals the outcomes of each approach. While directly eliminating keypoints based on uncertainty seems straightforward, it risks excessive removal, potentially undermining the process. On the other hand, ranking keypoints by uncertainty offers a more balanced and reliable choice.


\section{Conclusion}
\label{sec:conclusion}

We have introduced a new task of short-term 6-DoF object pose tracking in dynamic settings from RGB-only input. 
We focus on a model-free setup, only requiring segmentation of the object in the initial frame.
To facilitate the development and evaluation of this task, we created a large-scale synthetic dataset, OmniPose6D, for both training and evaluation. We also developed an uncertainty-aware keypoint refinement network, advancing the accuracy and reliability of object tracking. On both synthetic and real datasets, our approach demonstrates state-of-the-art results compared to existing methods. 
However, our pipeline still builds upon existing successes in segmentation, keypoint tracking, and bundle adjustment modules.
Future work will focus on the investigation of a broader range of temporal window settings, exploring the usage of the proposed dataset on more baselines, improving the pipeline's speed and exploring more effective methods for keypoint sampling and selection, alongside the simultaneous refinement of keypoints and pose.


\bibliographystyle{IEEEtran}
\bibliography{main.bib}

\clearpage
\setcounter{page}{1}
\section*{APPENDIX}

\setcounter{section}{0}
\renewcommand{\thesection}{\Alph{section}}
\renewcommand{\thesubsection}{\alph{subsection})}

\section{Model Details}
\label{sec:model_detail}

\noindent {\bf Architecture.}
Our model employs CoTracker~\cite{karaev2023cotracker}'s feature extraction process, analyzing each frame in an 8-frame sequence at a 384×512 resolution. It uses a 2D CNN to downsample the image by a factor of 8, outputting 128-dimensional features. The architecture consists of a 7×7 convolution (stride 2), followed by eight 3×3 kernel residual blocks with instance normalization, and concluding with two convolutions using 3×3 and 1×1 kernels. In both the training and evaluation phases, the model undergoes 4 iterative updates, respectively.

\noindent {\bf Training.}
Each training sample involves densely sampling keypoints on the initial mask with a grid size of 3. Due to GPU memory constraints, we randomly select 500 tracks from these samples. Contrary to CoTracker, which assumes stationary track points or fragments for initial track continuation, our approach offers two initial track settings. The first uses the original CoTracker predictions as a starting point for refinement, and the second applies gaussian noise (mean=0, std=1 pixel) to the ground truth keypoint for initial track generation, followed by refinement. This dual-mode initialization allows us to exclusively focus on the refinement task.
We also load the pre-trained weights from CoTracker, as shown in Table~\ref{tab:data_source}.

\noindent {\bf Loss.} We employ a L1 loss $L_{keypoint}$ for keypoint prediction and a cross-entropy loss for visibility $L_{vis}$ and uncertainty $L_{uncert.}$ predictions, respectively. 

Following CoTracker~\cite{karaev2023cotracker}, the keypoint loss is as follows:
\begin{equation}
{{\cal L}_{keypoint}}(\hat P,P) = \sum\limits_{m = 1}^M {{\gamma ^{M - m}}} \left\| {{{\hat P}^{(m)}} - P} \right\|
\end{equation}

where the optimization iteration number $M$ is set to 4, $\gamma = 0.8$ discounts early transformer updates, $P$ contains the ground-truth trajectories, and $\hat P$ is the prediction. 

The final loss is determined as follows: 
\begin{equation}
L_{all} = w_{keypoint} \cdot L_{keypoint} + w_{vis} \cdot L_{vis} + w_{uncert.} \cdot L_{uncert.}
\end{equation}
where \( w_{keypoint} \), \( w_{vis} \), and \( w_{uncert.} \) are hyperparameters weighting the loss terms. We use \( w_{keypoint} = 1 \), \( w_{vis} = 5 \), and \( w_{uncert.} = 5 \) in our experiments.
 
\section{Baselines}
\label{sec:baseline_detail}

\noindent {\bf LoFTR~\cite{sun2021loftr}.} We adopt the LoFTR outdoor model, pre-trained on MegaDepth~\cite{li2018megadepth}, following the previous object pose estimation method~\cite{he2022onepose++}. Additionally, we include mask information to ensure a fair comparison.

\noindent {\bf DROID-SLAM~\cite{teed2021droid}.} We use masked RGB input and follow the default settings specified in the demonstration script available at the official repository, where initialization frames are set to 8.

\noindent {\bf TAPIR~\cite{doersch2023tapir}.} We employ the standard TAPIR model, which is claimed to have improved performance. The samples originate from the mask in the initial frame while masked RGB input is provided.

\noindent {\bf CoTracker~\cite{karaev2023cotracker}.} We operate with the default settings, using the pre-trained model with a stride of 4 and a window size of 8. Similar to others, it processes masked RGB input, with samples drawn from the mask in the initial frame.

\section{Benchmark datasets}

We assess the performance of baseline methods on the OmniPose6D Synthetic dataset and three real datasets: HO3D~\cite{hampali2020honnotate}, YCBInEOAT~\cite{wen2020se}, and DexYCB~\cite{chao2021dexycb}.
Each dataset offers distinct challenges, ranging from hand-object interactions to cluttered environments. The results are shown in Section~\ref{sec:exp} and Section~\ref{sec:exp_additional}. We evaluate the baselines without fine-tuning to gauge their out-of-the-box capabilities. We highlight the performance enhancements by comparing them against our pipeline built upon the best-performing model CoTracker with improved design and comprehensive OmniPose6D training data. This comparison not only underscores the benefits of our synthetic data and tailored model design in enhancing generalized capabilities across diverse scenarios and targets but also demonstrates our approach's effectiveness in bridging the simulation-to-reality gap.

\section{Pipeline Details}
\label{sec:pipeline_detail}

As outlined in Section~\ref{exp:synthetic}, we employ a keypoint tracker~\cite{karaev2023cotracker} to determine keyframes. Specifically, we process input images using a sliding window approach with a window size of 8. For each window, a segmentation mask is applied to the first frame, and we set the grid size to 50 to generate sparse keypoints. The average motion of these sampled keypoints is then estimated. A keypoint is considered to have sufficient lateral movement if the L2 distance between its tracks in two adjacent views exceeds a defined threshold, $t_{dis}=2.0$. Similarly, if the L2 distance between a pair of keypoints in two adjacent views changes beyond $t_{scale}=1.3$, it is indicative of substantial zoom-direction motion. This preprocessing step, though simple, is effective in filtering out frames where the object remains stationary. Such stationary frames could otherwise violate the assumptions underlying our Structure-from-Motion framework.
This keyframe strategy is also similar to the one adopted in DROID-SALM~\cite{teed2021droid} which is based on the optical flow threshold. Note that no Ground Truth information (\eg pose) is used for this preprocessing step.

\begin{figure*}[t]
\centering
\includegraphics[width=0.4\textwidth]{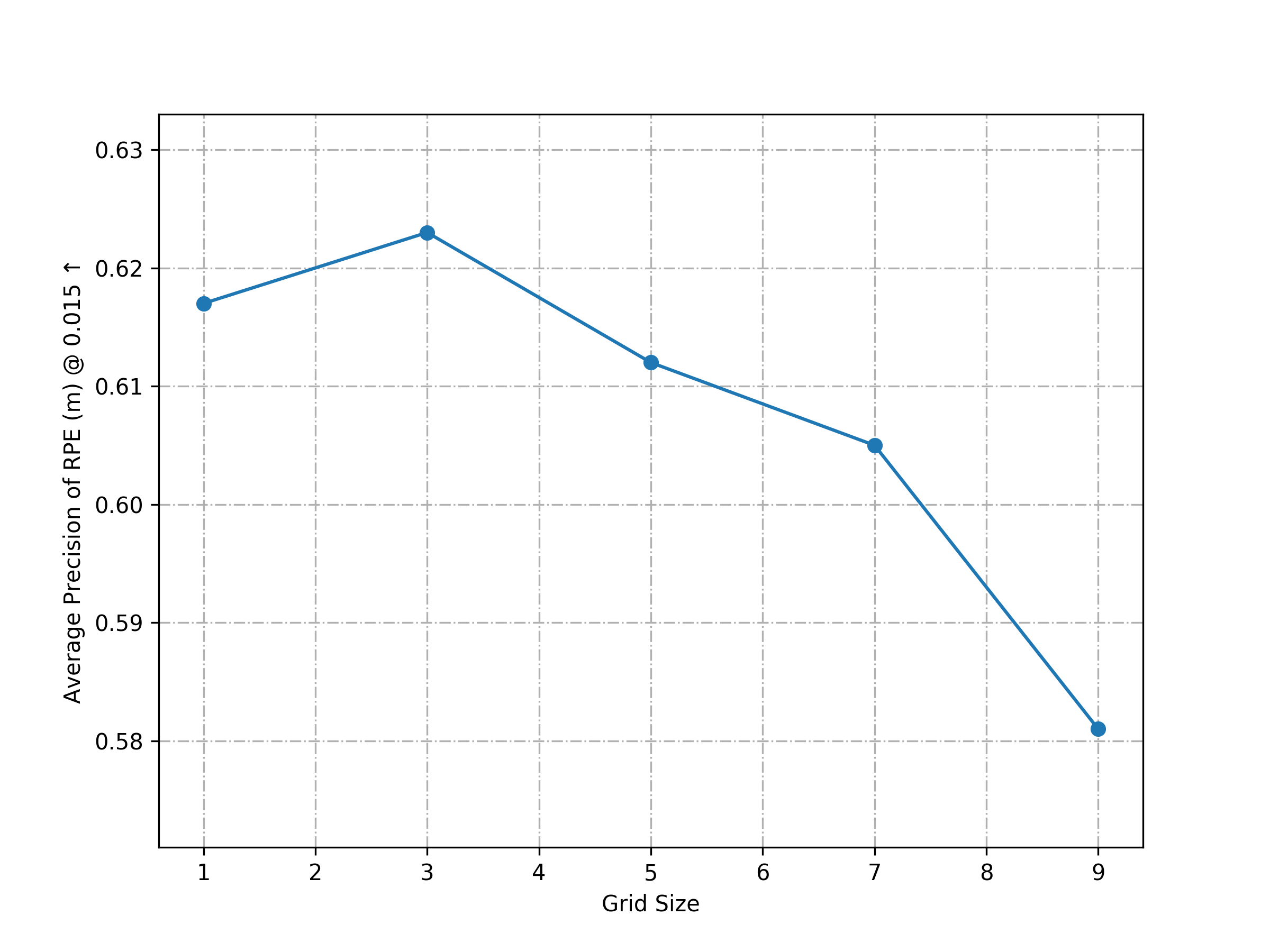} 
\includegraphics[width=0.4\textwidth]{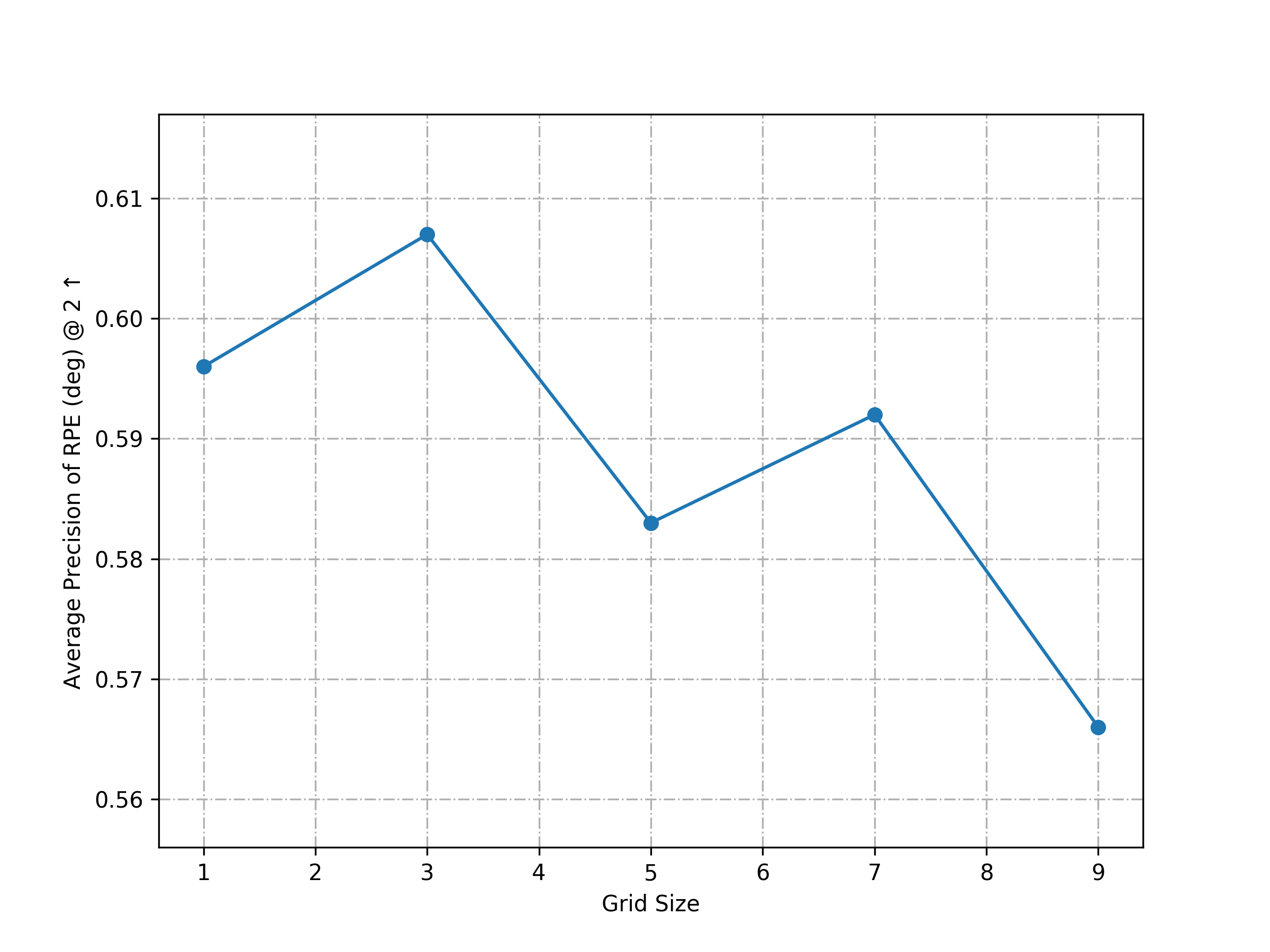} 
\caption{{\bf Ablate on Sampling Grid Size on the HO3D~\cite{hampali2020honnotate} dataset.} We ablate the grid size of the sampling on the mask in the first frame during inference. It is important to note that a grid size of 3 was employed during the training phase. \label{fig:ablate_grid_Size}}
\end{figure*}

\begin{figure*} [t]
\centering
\includegraphics[width=0.4\textwidth]{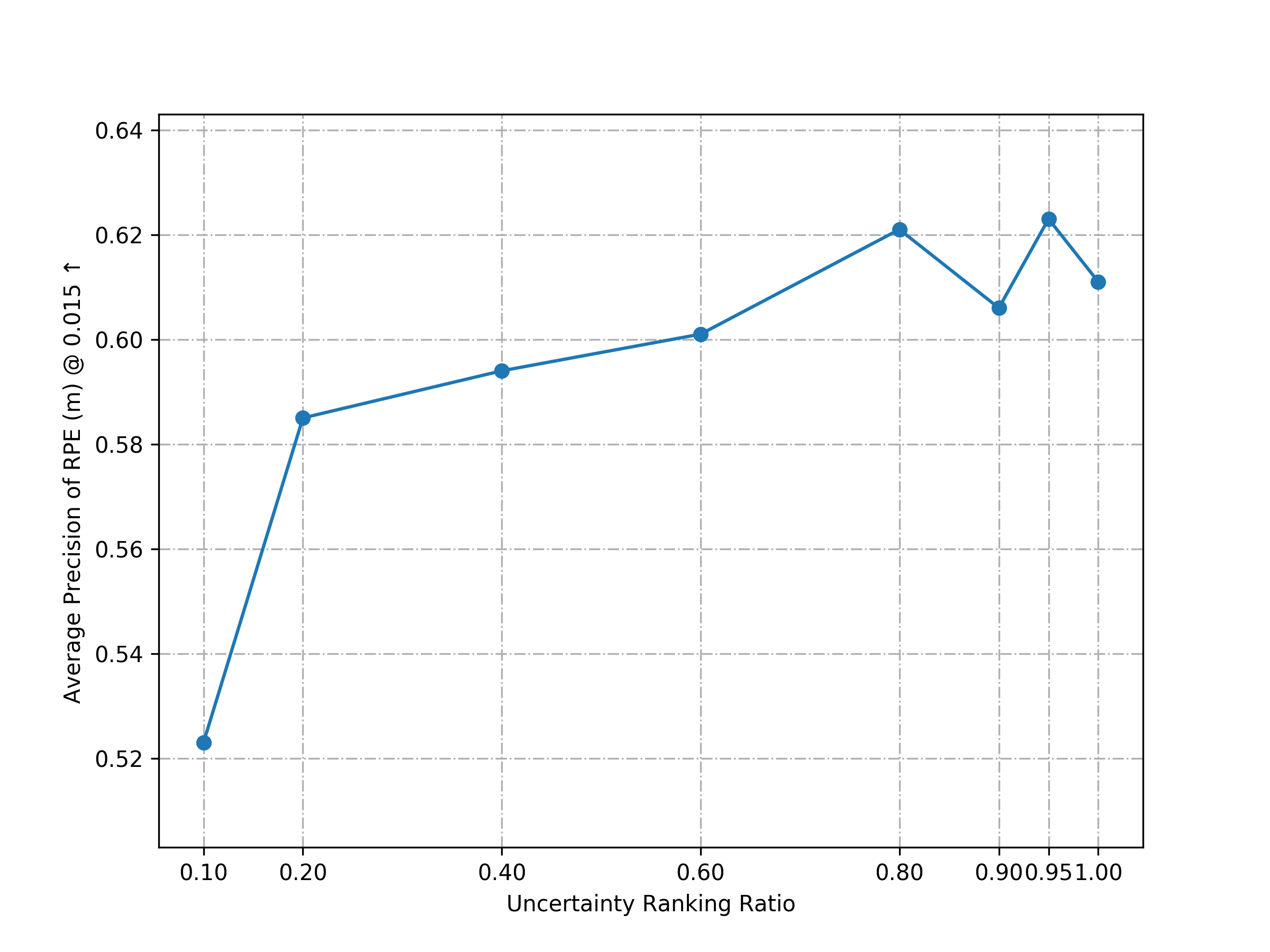} %
\includegraphics[width=0.4\textwidth]{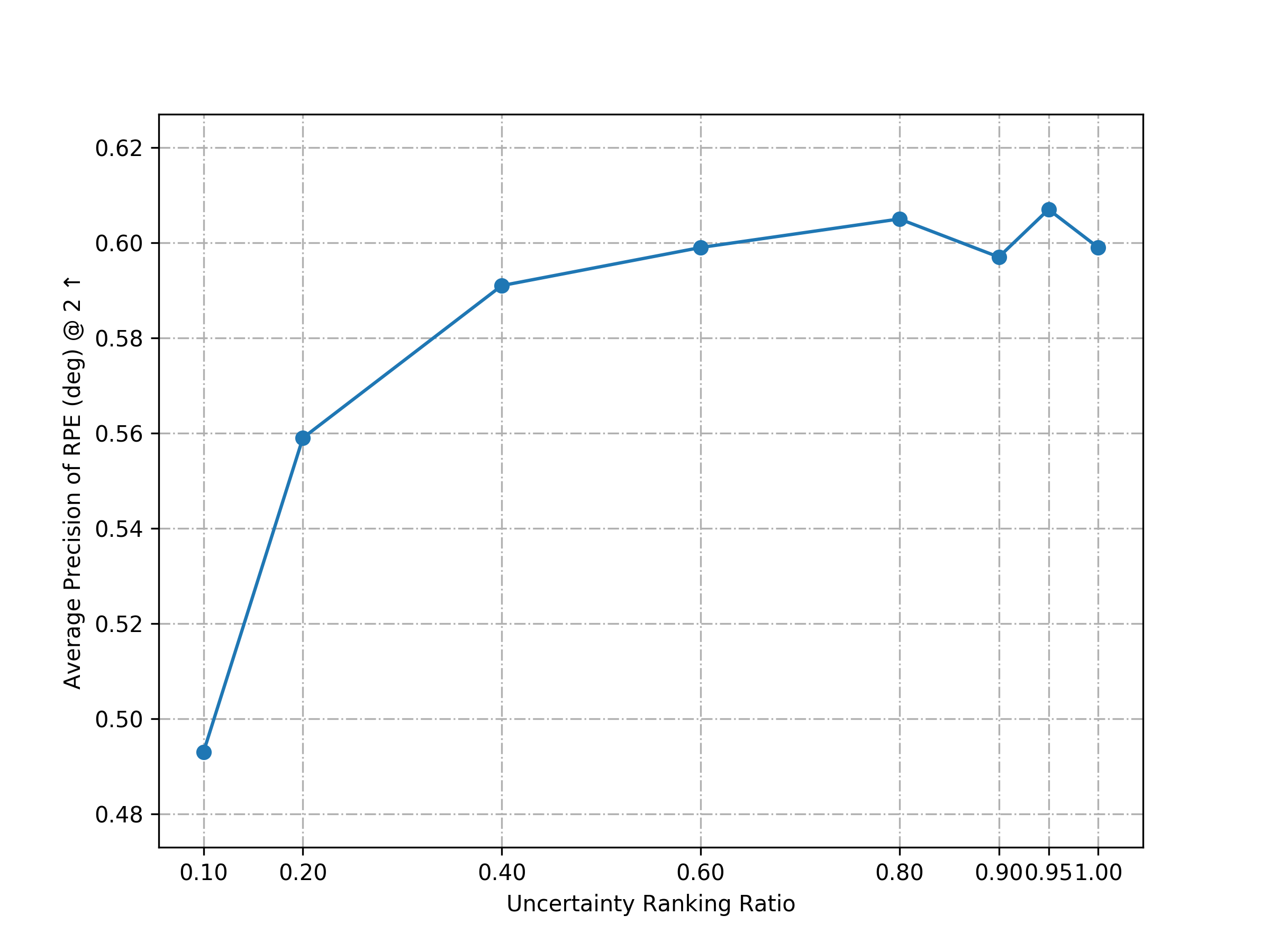} 
\caption{{\bf Ablate on Uncertainty Hyper-parameter Setting on the HO3D~\cite{hampali2020honnotate} dataset.} We ablate the ratio hyper-parameter when using the ranking strategy in uncertainty estimation. \label{fig:ablate_uncertainty}}
\end{figure*}

\section{More Ablation Studies}
\label{sec:more_ablation}

We ablate our design choices on HO3D~\cite{hampali2020honnotate} given its more realistic settings.

\noindent {\bf Sample Point Grid Size.} We ablate the grid size of sampling on the mask in the first frame during inference. Figure~\ref{fig:ablate_grid_Size} depicts the average precision of Relative Pose Error (RPE) at 0.015m and Relative Pose Error (RPE) at 2 degrees, respectively. We observe that the performance does not increase with a smaller grid size. This is attributed to our limited GPU memory, which necessitates random selection of points from the samples. Consequently, this may lead to insufficient sampling in certain areas. On the other hand, performance declines with a larger grid size (greater than 3), which is likely due to a grid size of 3 during training, and the attention module inherently encoding point relationships within the network. Thus, we choose to keep the same grid size setting during both training and inference for optimal results. 

\noindent {\bf Uncertainty Hyperparameter Setting.}
We further explore the impact of the ratio hyperparameter in our uncertainty estimation's ranking strategy. Figure~\ref{fig:ablate_uncertainty} depicts the average precision of Relative Pose Error (RPE) at 0.015m and Relative Pose Error (RPE) at 2 degrees, respectively. Our observations reveal a significant performance decline with more stringent settings (i.e., retaining far fewer points). In contrast, a ratio setting greater than 0.8 yields improved results.
This phenomenon could be attributed to the challenges inherent in accurately estimating the uncertainty of keypoints. Allowing a less restrictive threshold, which retains more candidates while filtering out the most unreliable ones, may be more effective. In this context, the Structure-from-Motion (SfM) framework can further refine poses during bundle adjustment. Based on these considerations, we empirically set the ratio to 0.95 to achieve optimal results.

\begin{figure*}[t!]
  \centering
  \vspace{1mm}

  \begin{minipage}{0.6\textwidth}
    \centering
    \includegraphics[width=\textwidth]{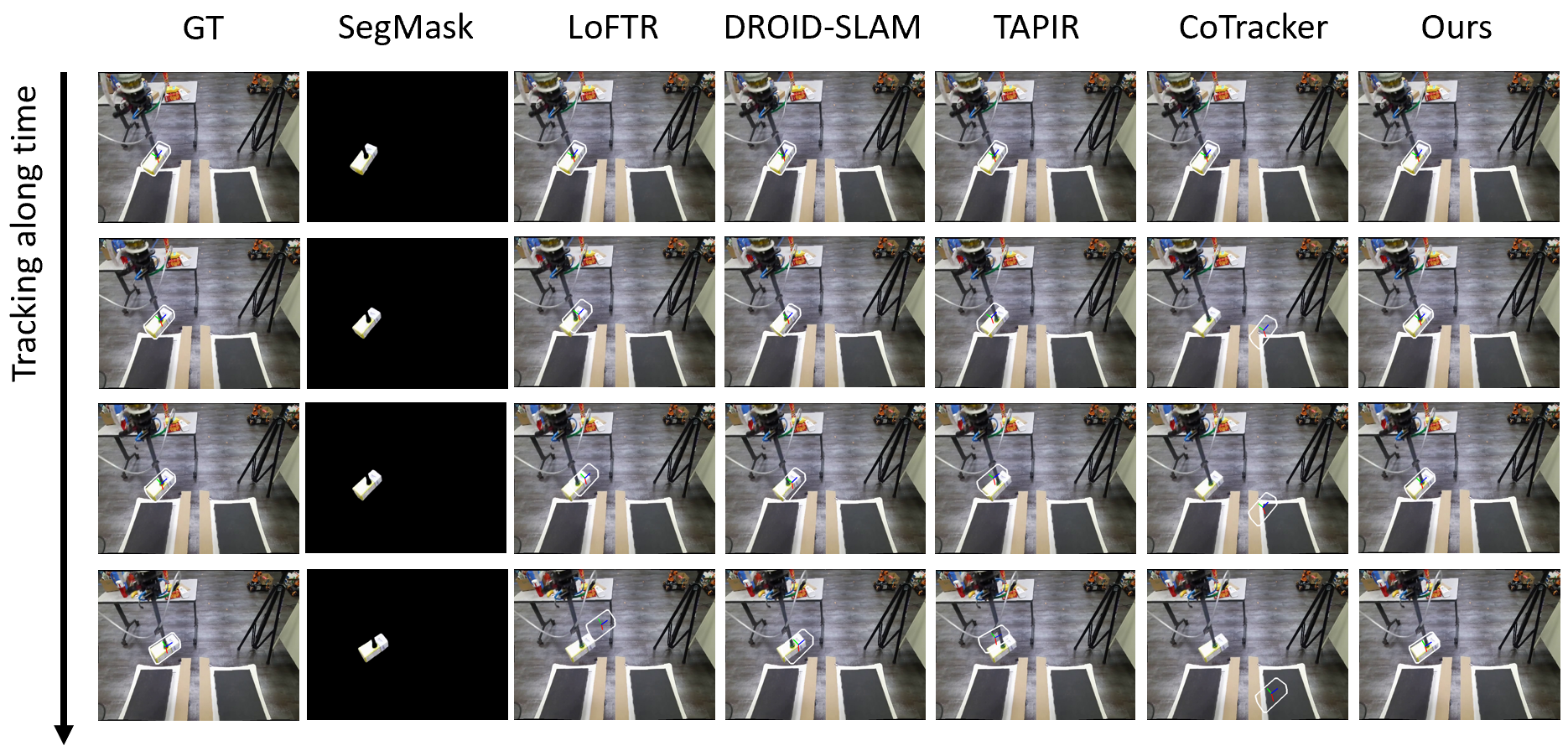}
    \vspace{-6mm} \\ 
    {\small (a) YCBInEOAT dataset}
  \end{minipage}

  \vspace{1mm}

  \begin{minipage}{0.6\textwidth}
    \centering
    \includegraphics[width=\textwidth]{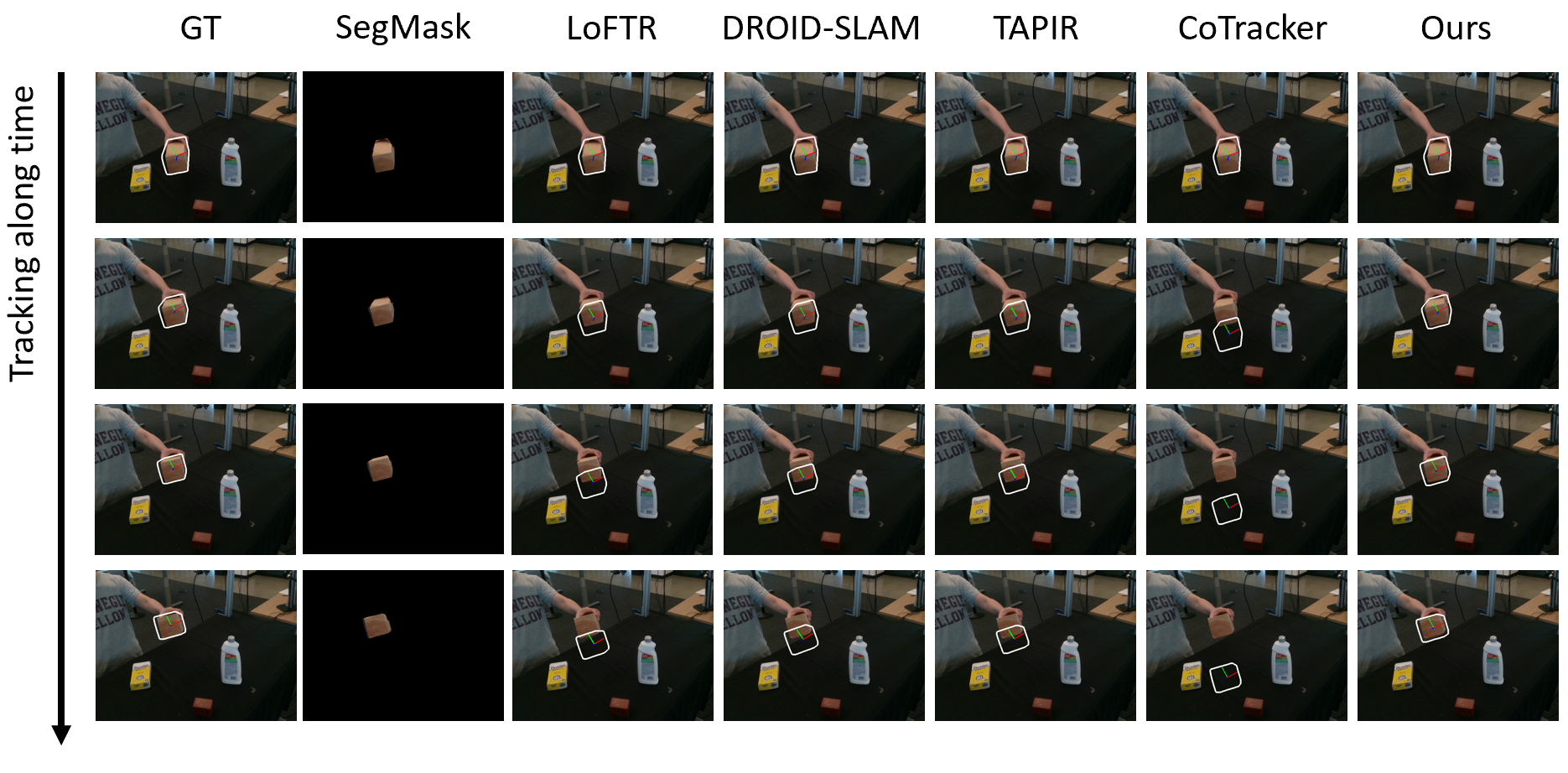}
    \vspace{-6mm} \\
    {\small (b) DexYCB dataset}
  \end{minipage}

  \caption{Qualitative comparison of the methods on the YCBInEOAT dataset~\cite{wen2020se} and DexYCB dataset~\cite{chao2021dexycb}. We visualize 6-DoF pose tracking results, where the contour (white) is rendered with the estimated pose. Only our proposed method successfully tracks the object.}
  \label{fig:qualitative_comparison_combined}
\end{figure*}

\section{Additional Experiments on Real-world benchmarks}
\label{sec:exp_additional}

\begin{table*}[t]
\caption{Evaluation of methods on YCBInEOAT~\cite{wen2020se} and DexYCB~\cite{chao2021dexycb} datasets with KITTI Odometry metrics. We report average precision ($\uparrow$) for each metric, conditional on the error below several thresholds.}
  \centering
  \begin{threeparttable}
  \resizebox{0.9\linewidth}{!}{
\begin{tabular}{cccccccccccccccc}
\toprule
\multirow{2}{*}{Method} & \multicolumn{3}{c}{$t_{err}$ (\%)} & \multicolumn{3}{c}{$r_{err}$ (deg/m)} & \multicolumn{3}{c}{ATE (m)} & \multicolumn{3}{c}{RPE (m)} & \multicolumn{3}{c}{RPE (deg)} \\
\cmidrule(r){2-4} \cmidrule(r){5-7} \cmidrule(r){8-10} \cmidrule(r){11-13} \cmidrule(r){14-16}
                        & 10 & 20 & 30 & 30 & 40 & 50 & 0.01 & 0.03 & 0.05 & 0.01 & 0.015 & 0.025 & 1 & 2 & 3 \\
\midrule
\multicolumn{16}{c}{\textbf{YCBInEOAT Dataset}} \\
\midrule
GT              & 0.33 & 0.51 & 0.56 & 0.54 & 0.57 & 0.59 & 0.51 & 0.71 & 0.73 & 0.62 & 0.69 & 0.73 & 0.55 & 0.65 & 0.69 \\
LoFTR~\cite{sun2021loftr}               & 0.00 & 0.00 & 0.01 & 0.00 & 0.02 & 0.04 & 0.02 & 0.25 & 0.43 & 0.06 & 0.16 & 0.35 & 0.05 & 0.15 & 0.33 \\
DROID-SLAM~\cite{teed2021droid}          & 0.03 & 0.03 & 0.03 & 0.03 & 0.04 & 0.04 & 0.07 & 0.33 & 0.52 & 0.17 & 0.29 & 0.49 & \textbf{0.15} & \textbf{0.34} & \textbf{0.58} \\
TAPIR~\cite{doersch2023tapir}               & 0.03 & 0.03 & 0.03 & 0.03 & 0.04 & \underline{0.07} & 0.05 & \underline{0.39} & \underline{0.65} & \underline{0.17} & \underline{0.30} & \underline{0.55} & 0.01 & 0.15 & 0.34 \\
CoTracker~\cite{karaev2023cotracker}           & \underline{0.03} & \underline{0.03} & \underline{0.03} & \underline{0.03} & \underline{0.03} & 0.06 & \underline{0.08} & 0.25 & 0.49 & 0.14 & 0.25 & 0.50 & 0.00 & 0.19 & 0.33 \\
Ours                & \textbf{0.03} & \textbf{0.03} & \textbf{0.04} & \textbf{0.04} & \textbf{0.08} & \textbf{0.13} & \textbf{0.08} & \textbf{0.39} & \textbf{0.67} & \textbf{0.17} & \textbf{0.32} & \textbf{0.57} & \underline{0.05} & \underline{0.24} & \underline{0.47} \\
\midrule
\multicolumn{16}{c}{\textbf{DexYCB Dataset}} \\
\midrule
GT              & 0.53 & 0.66 & 0.70 & 0.70 & 0.71 & 0.73 & 0.42 & 0.73 & 0.77 & 0.63 & 0.72 & 0.76 & 0.66 & 0.74 & 0.77 \\
LoFTR~\cite{sun2021loftr}               & 0.02 & \underline{0.09} & \underline{0.19} & 0.17 & 0.21 & 0.26 & \underline{0.01} & \underline{0.13} & \underline{0.22} & \underline{0.03} & \underline{0.10} & 0.22 & 0.03 & 0.18 & 0.28 \\
DROID-SLAM~\cite{teed2021droid}          & 0.00 & 0.04 & 0.13 & 0.13 & 0.19 & 0.26 & 0.00 & 0.08 & 0.17 & 0.02 & 0.05 & 0.18 & 0.02 & 0.20 & \underline{0.45} \\
TAPIR~\cite{doersch2023tapir}               & 0.00 & 0.06 & 0.16 & 0.16 & 0.24 & \underline{0.32} & 0.00 & 0.08 & 0.19 & 0.01 & 0.08 & 0.18 & 0.03 & 0.17 & 0.31 \\
CoTracker~\cite{karaev2023cotracker}           & 0.00 & 0.04 & 0.15 & \underline{0.17} & \underline{0.25} & 0.29 & 0.00 & 0.10 & 0.19 & 0.01 & 0.08 & \underline{0.22} & \underline{0.03} & \underline{0.21} & 0.31 \\
Ours                & \textbf{0.02} & \textbf{0.17} & \textbf{0.29} & \textbf{0.27} & \textbf{0.40} & \textbf{0.47} & \textbf{0.01} & \textbf{0.24} & \textbf{0.38} & \textbf{0.06} & \textbf{0.18} & \textbf{0.40} & \textbf{0.09} & \textbf{0.34} & \textbf{0.49} \\
\bottomrule
\end{tabular}
}
  \end{threeparttable}
\label{tab:main_combined}
  \vspace*{-4mm}
\end{table*}

\subsection*{a) YCBInEOAT Real Dataset}
\label{exp:YCBInEOAT}
\noindent{\bf Setting.} 
The YCBInEOAT~\cite{wen2020se} dataset contains object interactions with a dual-arm robot, captured by an Azure Kinect, with three distinct manipulation tasks: single-arm pick-and-place, intricate within-hand manipulation, and a seamless pick-to-hand-off transition between arms leading to placement. The dataset contains 5 diverse YCB objects, spanning 9 videos, each approximately 1000 frames long. We preprocess them in a similar way as detailed in Section~\ref{exp:ho3d}.

\noindent {\bf Results.}
We find our proposed method consistently achieves the best average performance (Table~\ref{tab:main_combined} (top)). However, a noticeable performance decline is observed compared to the results on the HO3D dataset~\cite{hampali2020honnotate}, except for DROID-SLAM~\cite{teed2021droid}. This disparity is likely due to the intrinsic characteristics of the datasets. 
Contrary to HO3D, where the hand rotates the object within a short-term window, the manipulator in the YCBInEOAT dataset always grasps the object, moving in a straight line with minimal rotational change.
This limited movement pattern challenges the Structure-from-Motion framework's assumptions, potentially complicating the accurate computation of poses. On the other hand, DROID-SLAM's approach of jointly updating keypoints and poses appears more robust against these challenges, encouraging future exploration of more integrated designs.
Furthermore, we presented a case study of robot grasping within the YCBInEOAT dataset (Figure~\ref{fig:qualitative_comparison_combined} (top)), where a robotic suction device is used to rotate the target object for subsequent manipulation tasks. Overexposure to light is a prevalent issue that hampers the performance of most baseline methods in tracking the target. However, our approach remains effective under such conditions. It's important to note that although our methodology builds upon CoTracker, this foundation showed limitations due to its noisy keypoint tracking results. Our method, on the other hand, benefits from explicitly calculating the uncertainty of each keypoint and removes the unreliable ones.
\subsection*{b) DexYCB Real Dataset}
\label{exp:DexYCB}
\noindent {\bf Setting.}
The DexYCB dataset~\cite{chao2021dexycb} captures hand graspings of 20 YCB objects by 10 subjects, under 8 mid-to-far positioned RGB-D cameras (RealSense D415).
Given our primary interest lies in the interaction with objects rather than the subjects themselves, we opted for the unseen grasping (S3) test split, selecting all the videos with subject 0. This subset comprises 120 videos, each containing 72 frames. We followed a similar pre-processing procedure detailed in Section~\ref{exp:ho3d}.    

\begin{figure}[t!]
  \centering
    \includegraphics[width=\linewidth]{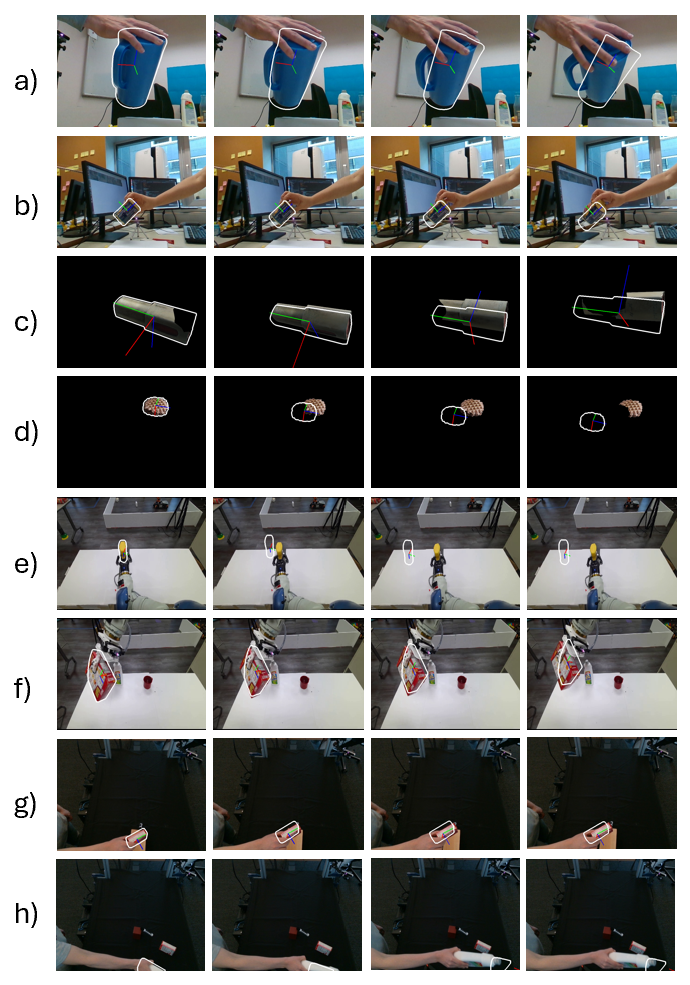}
  \caption{Representative failure cases of our proposed method on the  OmniPose6D (a and b); HO3D~\cite{hampali2020honnotate} (c and d);
  YCBInEOAT~\cite{wen2020se} (e and f);
  DexYCB dataset~\cite{chao2021dexycb} (g and h). Challenges include textureless regions, coplanar points, out-of-plane object rotation, small object translation, etc.}
  \label{fig:failure_case}
  \vspace{-5mm}
\end{figure}

\noindent {\bf Results.}
We observe our method outperforms others across all evaluated metrics (Table~\ref{tab:main_combined} (bottom)). While the average performance on this dataset does not match that achieved on the HO3D dataset, we hypothesizs this discrepancy can be attributed to the DexYCB dataset's emphasis on object-picking scenarios, characterized by significant hand occlusion and cases where the object is positioned far from the camera, leaving limited areas visible.
This challenge unveils new opportunities for research in the domain of joint hand and object pose estimation.
Additionally, we showcased a human grasping sample (Figure~\ref{fig:qualitative_comparison_combined} (bottom)) within the DexYCB dataset~\cite{chao2021dexycb}, where the object of interest is a textureless wooden cuboid. This scenario often poses a significant challenge for most baseline methods due to their inability to effectively process limited visual information. In contrast, our approach excels by capitalizing on a more extensive training dataset, which includes a broader spectrum of meshes and scenarios.

\section{OmniPose6D Dataset Sample Videos}

We showcased the varied range of high-fidelity meshes and motion trajectories within the OmniPose6D dataset. For detailed insights, please refer to the accompanying video supplements.

\section{Limitations}
\label{sec:limit}

Our proposed tracking method outperforms existing ones, yet achieving robust tracking in dynamic settings remains a challenging endeavor.
We highlight this through representative failure cases illustrated in Figure~\ref{fig:failure_case}. The encountered challenges are often presented as a blend of the following:
{\bf 1)} Difficulty in accurate pose tracking in textureless regions;
{\bf 2)} Insufficient geometric information for pose computation in the Structure-from-Motion framework when initial keypoints are predominantly sampled on a single surface (coplanar);
{\bf 3)} Challenges where limited texture information and large out-of-plane rotations of objects occur;
{\bf 4)} Scenarios where small translational movements of objects disrupt the assumptions of Structure-from-Motion, making it challenging to initialize a relative pose accurately.
These examples highlight the complexity of the task and provide essential guidance for future initiatives, such as creating a more comprehensive training dataset that includes corner cases.
In addition to that, the pipeline currently functions in an offline mode, and the keypoint selection process still needs further development. We will address these aspects in future work.

\end{document}